\documentclass[runningheads]{llncs}
\usepackage{graphicx}
\usepackage{xcolor}
\usepackage{amsmath}
\usepackage{amsfonts}
\usepackage{subcaption}
\usepackage{algorithm}
\usepackage{algorithmic}
\usepackage{multirow}
\usepackage{booktabs}
\usepackage{svg}

\usepackage[hidelinks]{hyperref}
\begin{document}
%
\title{Improved Counting and Localization from Density Maps for Object Detection in 2D and 3D Microscopy Imaging}
%
%
\author{Shijie Li\inst{1} \and
Thomas Ach\inst{2} \and
Guido Gerig\inst{1}}
%
\authorrunning{S. Li et al.}
%
\institute{ Computer Science and Engineering, New York University, Brooklyn, NY, USA. \email{\{shijie.li, gerig\}@nyu.edu} \and
Ophthalmology, University Hospital Bonn, Bonn, Germany}
%
\maketitle              
\begin{abstract}

Object counting and localization are key steps for quantitative analysis in large-scale microscopy applications. This procedure becomes challenging when target objects are overlapping, are densely clustered, and/or present fuzzy boundaries. Previous methods producing density maps based on deep learning have reached a high level of accuracy for object counting by assuming that object counting is equivalent to the integration of the density map. However, this model fails when objects show significant overlap regarding accurate localization. We propose an alternative method to count and localize objects from the density map to overcome this limitation. Our procedure includes the following three key aspects: 1) Proposing a new counting method based on the statistical properties of the density map, 2) optimizing the counting results for those objects which are well-detected based on the proposed counting method, and 3) improving localization of poorly detected objects using the proposed counting method as prior information. Validation includes processing of microscopy data with known ground truth and comparison with other models that use conventional processing of the density map. Our results show improved performance in counting and localization of objects in 2D and 3D microscopy data. Furthermore, the proposed method is generic, considering various applications that rely on the density map approach. Our code will be released post-review.

\end{abstract}
\begin{figure}
    \centering
    \includegraphics[width=0.9\textwidth]{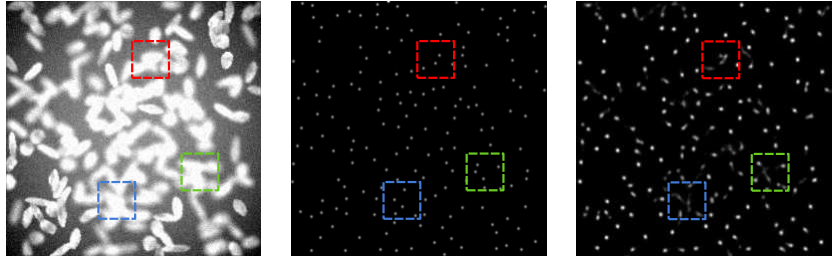}
    \caption{\textbf{Left to right}: the input image, its "ground truth" density map, and the predicted map produced by~\cite{ref_saunet}. The areas highlighted by the {\color[rgb]{1,0,0} red}, {\color[rgb]{0.4,0.8,0.13}green} and {\color[rgb]{0.25,0.48,0.84} blue} boxes contain overlapping objects which typically cause localization problems.}
\end{figure}
\section{Introduction\label{section:intro}}
Object counting and localization techniques are important processing steps for highly automated quantitative analysis of large-scale microscopy images driven by biological and medical applications. The main aim is the prediction of the number of target objects such as cells, nuclei, granules and also accurate localization of these objects for subsequent semantic segmentation. Such tasks are often based on simple point annotations, which take significantly less effort compared to precise outlining of sought objects. The use of point annotation becomes particularly important when objects are densely clustered, but also for object analysis from 3D imagery where manual outlining may not be feasible.

Supervised object counting and localization methods fall into two categories: counting~\cite{ref_saunet,vgg,bayesian,countcept,generalized,fcrn,multicolumn} and detection emphasized~\cite{celldirection,locality,xie2015beyond}. The former faces limitations with overlapping objects in regard to counting performance. Modern crowd counting methods are based on the training of a regression method that maps images to density maps, the integration of which is shown to be equivalent to the number of crowd objects. Placing a Gaussian kernel for each annotated dot is the most common approach to generate the density map. Lempitsky et al.~\cite{vgg} train a linear transform from the feature space to the density map space. Others~\cite{multicolumn,csrnet,fcrn} proposed convolutional neural networks as  feature extractors. Cohen et al.~\cite{countcept} generate the density map with a square kernel instead of a Gaussian kernel. Improving the objective function for optimization was proposed by~\cite{bayesian,generalized}. Recent work by Guo et al.~\cite{ref_saunet} shows superior counting performance on various cell datasets. To localize the objects, they use the connected components after thresholding or take local maxima. However, the evaluation of localization performance in ~\cite{ref_saunet} indicates that these postprocessing steps fail to predict the center of the object and tend to produce false positives when occlusion occurs, and do not seem to be robust enough to background noise. Furthermore, those methods are sensitive to hyperparameters such as threshold values.

In this paper, we propose a counting method that automatically determines the threshold to estimate the number of objects in a connected component. We use Gaussian mixture fitting to decode this information from the predicted density map. These two methods result in a postprocessing approach of the density map without the need for any additional hyperparameters. We validate our approach on multiple 2D and 3D datasets and demonstrate that the proposed method provides improved density maps and increases counting and localization performance in comparison to previous baselines.

\section{Density Map Analysis\label{section:anlyz}}

\noindent\textbf{Overview}
In this section, we briefly summarize the basic idea of the density map (DM) approach. We then provide a mathematical analysis of the issues of density maps generated with Gaussian kernels. Density map methods assume that $N$ images $\mathbf{I}_1, \mathbf{I}_2, \dots, \mathbf{I}_N \in \mathbb{R}^{H\times W\times C}$ and their corresponding annotations which are N sets of 2D/3D points $P_1, P_2, \dots, P_N$ are provided. For each point annotation set $P_i$ we generate a density map $\mathbf{D}_i\in \mathbb{R}^{H\times W}_{>0}$ as follows,

\begin{equation}
    \mathbf{D}_i = \sum_{j=1}^{C_i}\mathbf{D}_i^j, \quad
    \mathbf{D}_i^j(\mathbf{x}) = \mathcal{N}(\mathbf{x}; \mathbf{p}_i^j, \mathbf{\Sigma}) \label{eq:gaussiankernel}
\end{equation}
where $C_i\in \mathbb{Z}$ is the number of objects in the image $\mathbf{I}_i$, $\mathbf{p}_i^j\in P_i$, $\mathbf{x}$ is the pixel location, and $\mathbf{\Sigma}$ is the covariance matrix often defined related to the object size and the image isotropy. The learning step of the DM method is a training of a regressor mapping from the image $\mathbf{I}_i$ to the density map $\mathbf{D}_i$. During testing, the counting prediction $\hat{C}_i=\sum_{\mathbf{x}}\mathbf{\hat{D}}_{i}$ is computed, where $\mathbf{\hat{D}}_i$ is the predicted density. \\

\noindent\textbf{Decomposition of the Density Map}
Localizing of objects from the density map is based on a decomposition of the predicted density map by solving,
\begin{equation}
\min \quad \|\hat{\mathbf{D}}_i - \sum_{j=1}^{\hat{C}_i}\hat{\mathbf{D}}_i^j\|,\quad \textrm{s.t.} \quad \sum_{\mathbf{x}}\hat{\mathbf{D}}_i^j=1.  
\label{eq:optprob1}
\end{equation}
Applying connected component analysis after thresholding the density map is one simple but effective way to solve this problem. However, it fails to take additional constraints into consideration, which might have the risk of undercounting several objects merged into one or overcount when assigning background noise to an object. Furthermore, a particular threshold value has to be tuned manually during the training stage individually for different datasets. Therefore, we are looking for a way to determine the threshold automatically and to estimate the number of objects in connected components. We make the following assumptions:
\begin{enumerate}
    \item\label{asm:1} A thresholding perfectly decomposes the density map. That is to say, \\
    \noindent\begin{minipage}{.5\linewidth}
    \begin{equation}
        \mathbf{1}_{\hat{\mathbf{D}}_i>T} = \sum_{j=1}^{\hat{C}_i}\mathbf{1}_{\hat{\mathbf{D}}_i^j>T}\label{eq:thres},
    \end{equation}
    \end{minipage}
    \noindent\begin{minipage}{.5\linewidth}
    \begin{equation}
    \hat{\mathbf{D}}_i = \sum_{j=1}^{\hat{C}_i}\hat{\mathbf{D}}_i^j\label{eq:densitysum}.
    \end{equation}
    \end{minipage}
    \item\label{asm:2} Each decomposed component is the same as Eq. \ref{eq:gaussiankernel} with the same $\mathbf{\Sigma}$.
\end{enumerate}
Multiplying Eq. \ref{eq:thres} with Eq. \ref{eq:densitysum} , we then get
$
    \mathbf{1}_{\hat{\mathbf{D}}_i>T}\hat{\mathbf{D}}_i = \sum_{j=1}^{\hat{C}_i}\mathbf{1}_{\hat{\mathbf{D}}_i^j>T}\sum_{j=1}^{\hat{C}_i}\hat{\mathbf{D}}_i^j
    \label{eq:multiply}
$. According to Assumption \ref{asm:1}, we can have $\mathbf{1}_{\hat{\mathbf{D}}_i^j>T}\hat{\mathbf{D}}_i^k\approx\mathbf{0}$ if $j\neq k$. Eq.~\ref{eq:multiply} can then be written as 
$
    \mathbf{1}_{\hat{\mathbf{D}}_i>T}\hat{\mathbf{D}}_i = \sum_{j=1}^{\hat{C}_i}\mathbf{1}_{\hat{\mathbf{D}}_i^j>T}\hat{\mathbf{D}}_i^j
$. Following Assumption \ref{asm:2}, we can get $\sum_{\mathbf{x}}\mathbf{1}_{\hat{\mathbf{D}}_i^j>T}\hat{\mathbf{D}}_i^j=F(r_T)$, where $r_T$ is the Mahalanobis distance satisfying $\mathcal{N}(r_T, \mathbf{p_i^j}, \mathbf{\Sigma})=T$. Then the optimization problem Eq.~\ref{eq:optprob1} is equivalent to
    \begin{align}
    \min \quad & \|\mathbf{1}_{\hat{\mathbf{D}}_i>T}\hat{\mathbf{D}}_i - \sum_{j=1}^{\hat{C}_i}\mathbf{1}_{\hat{\mathbf{D}}_i^j>T}\hat{\mathbf{D}}_i^j\|, \\
    \textrm{s.t.} \quad & \sum_{\mathbf{x}}\mathbf{1}_{\hat{\mathbf{D}}_i^j>T}\hat{\mathbf{D}}_i^j/F(r_T)=1.  \label{eq:optprob2_ct}
    \end{align}

This indicates that if we choose the right threshold, Eq.~\ref{eq:optprob2_ct} should hold.

\begin{figure}[t!]
  \centering
  \includegraphics[width=0.9\textwidth]{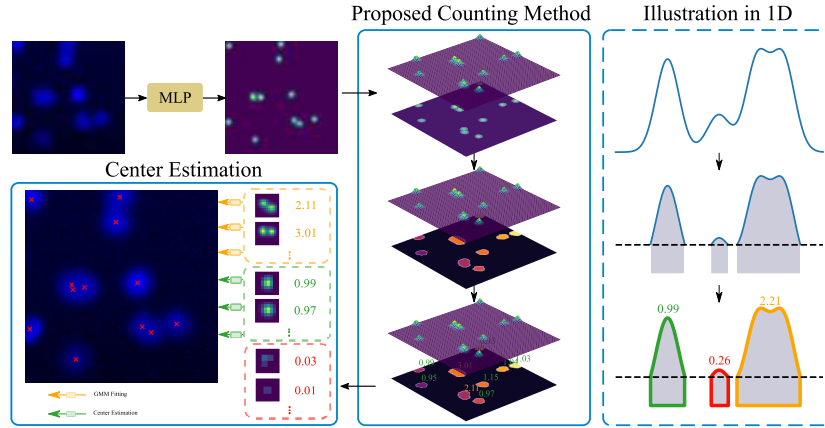}
  \caption{Overall pipeline: Our method contains two modules to be applied at the testing stage: 1) a counting method, and 2) center estimation. The threshold value will be determined automatically to identify the number of objects inside a connected component. The center estimation will localize the object center according to the number of objects in the connected component. Section~\ref{section:methodology} will discuss the proposed counting and localization procedures in detail.}
  \label{fig:pipeline}
\end{figure}

\section{Methodology \label{section:methodology}}
In the previous section, we reframed the localization problem into a modified density map decomposition problem when two assumptions hold. In real applications, there are two situations that may break these assumptions. For the convenience of illustration, we name the value on the left side of Eq.~\ref{eq:optprob2_ct} the count of the connected component.

\subsection{Regression Noise}
\noindent\textbf{Problem Statement}
While training the density map regressor, there is no constraint to assure Assumption 2 to hold. This means that even though we may choose the right threshold and the decomposition satisfies Assumption~\ref{asm:1}, the count of a connected component might be slightly greater or less than one. We call this "regression noise". In addition,  density maps may depict small peaks due to  background noise. \\
\noindent\textbf{Solution} We observe that the connected component count will be around 1 if a single object exists and 0 for no objects. A straightforward remedy would be to round the count of connected component between 0.5 and 1.5 to 1, for example. The connected components in green from Figure \ref{fig:pipeline} are identified as containing one object, while those in the red box no object.

\subsection{Imperfect Object Partition}
\noindent\textbf{Problem Statement}
In real cases, due to  clustering, overlapping objects and fuzzy edges, individual components in Eq.~\ref{eq:densitysum} are so close that it is hard to automatically find an optimal threshold, or a threshold for object separation may even not exist. In these situations, the Assumption~\ref{asm:1} does not hold.

\noindent\textbf{Solution}
There are two clues that will be used to approach problems with localization. For areas with multiple nearby objects, counting based on connected component  will give us a good counting estimate, and second, centers of objects will most likely show peaks in the density map. With these two observations, we a apply similar method as in \cite{gmmcurvefit} where we first normalize the density within the connected component area such that we can treat it as a probability distribution. We then perform Monte Carlo Sampling to generate random samples. Finally, we apply a Gaussian mixture model to fit the distribution. Centers of objects are then the fitted means of the Gaussian mixture.



\begin{table}[ht!]
  \centering
  \renewcommand{\arraystretch}{1.2}
  \begin{tabular}{cccccc}
    \toprule[1.5pt]
    \textbf{Dataset} & \textbf{Method} & \textbf{MAE}$(\downarrow)$ & \textbf{Precision}$(\uparrow)$ & \textbf{Recall}$(\uparrow)$ & \textbf{F-measure}$(\uparrow)$\\
    \midrule
    \multirow{5}{42pt}{\centering VGG} 
    & IoDM & $\mathbf{2.05(\pm 0.09)}$ & - & - & - \\
    & CCA-T & $17.18(\pm1.37)$ & $\mathbf{99.99(\pm 0.00)}$ & $91.20(\pm 0.47)$ & $95.36(\pm0.26)$ \\
    & DMA (ours) & $2.94(\pm0.47)$ & $98.32(\pm0.21)$ & $\mathbf{96.63(\pm0.24)}$ & $\mathbf{97.46(\pm0.18)}$ \\
    \midrule
    \multirow{3}{42pt}{\centering Ellipse} 
    & IoDM & $5.02(\pm 0.44)$ & $-$ & $-$ & $-$ \\
    & CCA-T & $6.48(\pm 0.96)$ & $\mathbf{98.74(\pm0.16)}$ & $95.77(\pm 0.5)$ & $97.20(\pm0.27)$ \\
    & DMA (ours) & $\mathbf{4.68(\pm 0.59)}$ & $98.15(\pm0.43)$ & $\mathbf{96.48(\pm0.48)}$ & $\mathbf{97.29(\pm0.30)}$\\
    \midrule
    \multirow{3}{42pt}{\centering MBM} & IoDM & $6.56(\pm 1.42)$ & - & - & - \\
    & CCA-T & $\mathbf{6.02(\pm1.47)}$ & $\mathbf{90.18(\pm0.86)}$ & $\mathbf{90.97(\pm1.30)}$ & $\mathbf{90.49(\pm0.69)}$ \\
    & DMA (ours) & $6.23(\pm1.12)$ & $89.55(\pm1.19)$ & $89.87(\pm1.63)$ & $89.63(\pm0.86)$ \\
    \midrule
    \multirow{3}{42pt}{\centering ADI} 
    & IoDM & $\mathbf{12.72(\pm1.75)}$ & - & - & - \\
    & CCA-T & $13.90(\pm 1.43)$ & $87.60(\pm1.02)$ & $90.47(\pm1.37)$ & $88.62(\pm1.11)$ \\
    & DMA (ours) & $13.45(\pm2.05)$ & $\mathbf{90.76(\pm1.30)}$ & $\mathbf{88.56(\pm1.69)}$ & $\mathbf{89.32(\pm1.19)}$\\
    \midrule
    \multirow{3}{42pt}{\centering DCC} 
    & IoDM & $3.12(\pm 0.39)$ & - & - & - \\
    & CCA-T & $3.17(\pm 0.53)$ & $\mathbf{90.75(\pm 1.11)}$ & $90.37(\pm 3.60)$ & $89.19(\pm 3.5)$ \\
    & DMA (ours) & $\mathbf{2.83(\pm0.30)}$ & $89.53(\pm2.05)$ & $\mathbf{90.53(\pm2.11)}$ & $\mathbf{89.28(\pm1.75)}$\\
    \midrule
    \multirow{3}{42pt}{\centering {BBBC027\\ (3D)}} 
    & IoDM & $\mathbf{13.58(\pm 4.54)}$ & - & - & - \\
    & CCA-T & $87.18(\pm 55.10)$ & $\mathbf{97.25(\pm 2.55)}$ & $79.25(\pm12.80)$ & $\mathbf{86.80(\pm10.00)}$ \\
    & DMA (ours) & $13.63(\pm 4.91)$ & $85.99(\pm 7.20)$ & $\mathbf{85.35(\pm6.23)}$ & $85.63(\pm6.70)$\\
    \midrule
    \multirow{3}{42pt}{\centering SIM (3D)} 
    & IoDM & $67.00(\pm 45.33)$ & - & - & - \\
    & CCA-T & $161.36(\pm 61.90)$ & - & - & - \\
    & DMA & $\mathbf{64.58(\pm 48.11)}$ & - & - & -\\
    \bottomrule[1.2pt]
  \end{tabular}
  \caption{Quantitative results of different methods on various test datasets. ($\uparrow$) means larger is better. $(\downarrow)$ means smaller is better.  \label{tab:synth}}
\end{table}

\begin{figure}[ht!]
    \centering
    \includegraphics[width=0.91\textwidth]{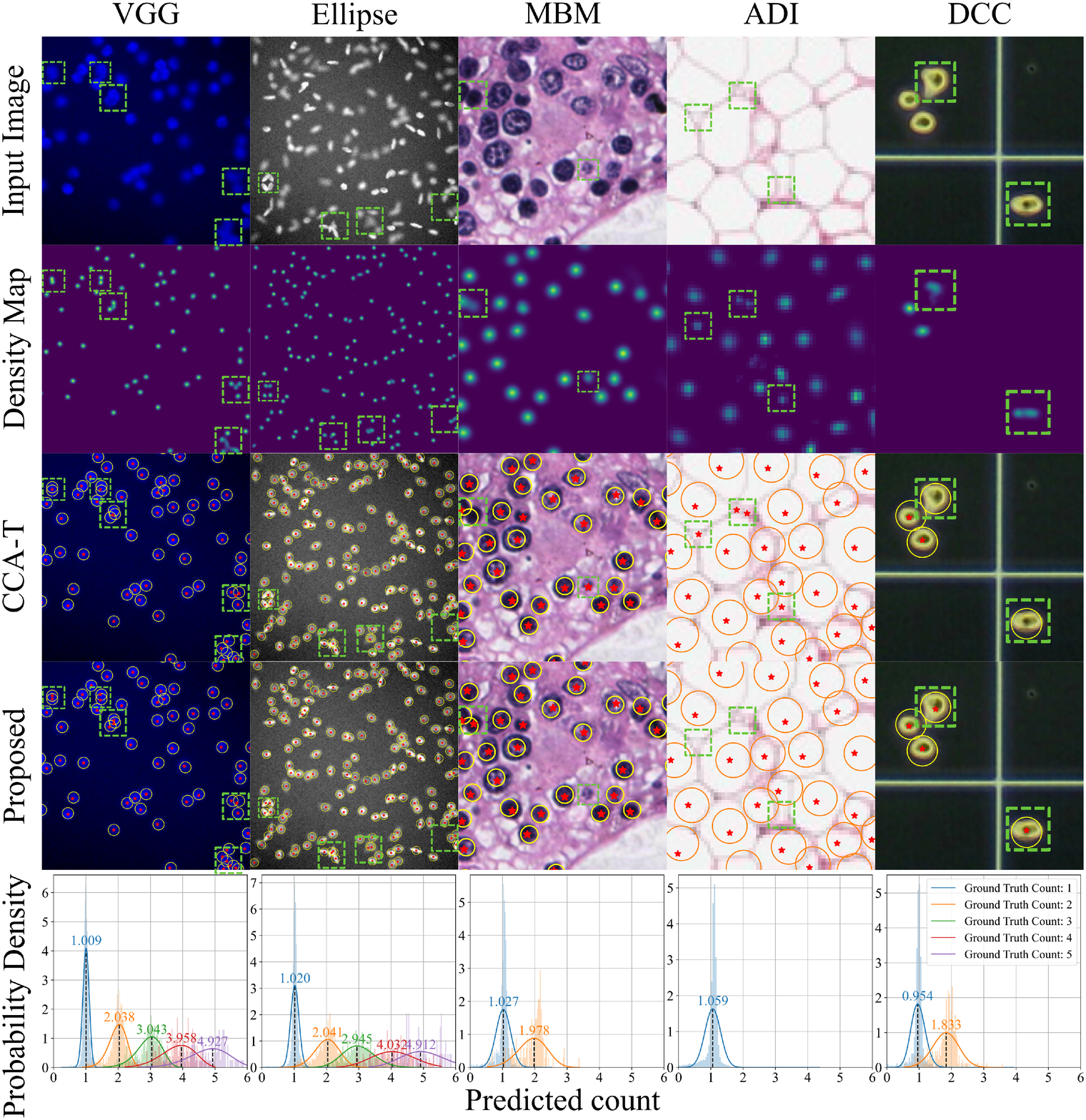}
    \caption{Illustration of the model's capability to localize objects while also coping with strongly overlapping structures, displayed for five different datasets (columns).   
    Circles represent the ground truth of object locations, and red stars mark the predicted object locations. Boxes mark locations where method's differ. Histograms (bottom row)  show posterior probability distributions of  predicted versus ground truth counts calculated from the testing split of the datasets. }
    \label{fig:qual}
\end{figure}
\section{Experiments}

Here, we introduce datasets used for verification, describe experimental configuration and evaluation metrics, and show qualitative and quantitative results. 
As we mentioned in sections \ref{section:intro} and \ref{section:anlyz}, commonly used methods to count and localize are integration of density map (IoDM) and connected component analysis with threshold (CCA-T). To evaluate  counting performance, we compare results of our proposed density map analysis (DMA) with those of the other methods when applied to different synthetic and real datasets.

\subsection{Datasets}
For a full comparison, all the 2D datasets as also used in~\cite{ref_saunet} are included for evaluation. Furthermore, we generate and use synthetic data where we can control occlusions, applying the tool from \cite{ref_simcep}.  We will also apply procedures to a volumetric Structural Illumination Microscopy (SIM) dataset to evaluate the method's performance in a clinical 3D application.

\noindent\textbf{VGG}
Simulated bacterial cells in fluorescence-light microscopy image using~\cite{ref_simcep}.
There are 200 $256\times256$ images with $176(\pm 61)$ cells in each image.\\
\noindent\textbf{Synthetic Ellipse Dataset}
By adding an eccentricity parameter to \cite{ref_simcep}, we generate a dataset containing object shape with eccentricities from $0$ to $0.85$. For each image, we allow a proportion $p$ of objects to overlap, where $p \sim U[0, 0.5]$.\\
\noindent\textbf{MBM}
The dataset was first introduced by \cite{YouShouldUseRegression} containing eleven $1200\times 1200$ images of healthy human bone marrow from eight different patients. The center point of the cell nuclei has been annotated. \cite{countcept} further crop it into 44 $600\times600$ images with $126\pm33$ cell nuclei in each image.\\
\noindent\textbf{DCC}
The Dublin Cell Counting datasets is quite challenging. It contains 177 images of stem cells from various tissues and species shot by different devices. Data therefore shows large variability in regard to the density, morphology, size and artifacts. The number of cells varies from 0 to 101.\\
\noindent\textbf{ADI}
We use images of human subcutaneous adipose tissue from \cite{GTEx}. \cite{countcept} cropped 200 regions of size $1700\times1700$ and downsampled them to $150\times150$. There are $165\pm 44$ cells on average in each image.\\
\noindent\textbf{BBBC027}
We used image set with low SNR in the BBBC027\cite{bbbc027} dataset for 3D validation. It synthesized 3D digital phantoms in human colon tissue. The weight center of the instance label is calculated as the point annotation.\\
\noindent\textbf{SIM}
RPE cells were imaged with super-resolution structured illumination microscopy (SIM) as part of a recent ophthalmological study \cite{bermond2020autofluorescent}. Our biomedical partners localized each organelle per cell with point annotations. There are 420 images in total. There are 424 objects on average in each image.

\subsection{Experimental setup  \label{section:setup}}

 We set up four experiments for each dataset. We first train with  similar steps as proposed in  \cite{ref_saunet}\footnote{The original tensorflow version is available at \url{https://github.com/mzlr/sau-net.git}. We reproduce results with pytorch.} and select the best models with the smallest counting mean square error based on IoDM and DMA separately. We then compare the counting performance with no postprocessing (IoDM), thresholding and connected-component analysis (CCA-T) and our density map analysis method (DMA). We find the best threshold of CCA-T during the validation step for each dataset. The image size, batch size, data augmentation, selection of the $\mathbf{\Sigma}$ in Eq. ~\ref{eq:gaussiankernel} and the scaling coefficient of the loss function are the same as \cite{ref_saunet}. To rule out randomness of results, we run 10 different data splits for each dataset and average the evaluation metrics.  All experiments are carried out on a desktop computer with an AMD Ryzen 9 5950X desktop processor and a single NVIDIA GeForce RTX 3080 GPU. Each experiment takes about one hour to complete.

\subsection{Evaluation Metrics \label{section:evalmetrics}}
We use Precision, Recall and F-measure to evaluate the model performance of localization, and MAE for the counting ability. A proposed object center is True Positive (TP) if and only if its distance to the nearest annotation points is less than a threshold (8 for MBM and 5 for the rest following~\cite{celldirection}). Each proposed point can only be assigned to one annotation point. Those annotation points without a corresponding proposal will be counted as False Negatives (FN). Proposals that are not True Positives are labeled as False Positives (FP).

\subsection{Results and Analysis \label{section:results}}
Figure~\ref{fig:qual} shows results for qualitative evaluation of our method on localization performance, highlighting that our method can distinguish strongly overlapping objects while also eliminating regression noise in those areas where the CCA-T method fails.
Table~\ref{tab:synth} lists quantitative results.  All methods present relatively high precision regarding localization among different datasets. But the proposed method generally has lower mean absolute counting error and higher recall and $F_1$ score especially in those datasets with severe object overlaps (VGG, Ellipse). This indicates that our method has better performance in counting and is producing less false positives of object localization in areas of clustered objects.

\section{Conclusion}
Crowd object counting and localization are important steps for quantitative analysis in large-scale microscopy applications. The consistency between counting and localization is also important for manual correction in certain applications where no ground-truth is available. In this work, we propose a counting method applied to deep network predicted density maps leveraging the property of multivariate normal distributions. Validation and comparison to previously proposed methods demonstrate improved performance in localization - an aspect which may also be important for subsequent semantic segmentation. Analysis also shows improvements in 3D data where there is rapidly growing interest for automated analysis.  Furthermore, our comparisons also reveal improvement of object counting results. The proposed processing is generic in view of crowd object counting applications based on density maps.  Future work will tackle the still existing problem of incorrect detection produced by the regression model. 
\bibliographystyle{splncs04}
\bibliography{refs}
\end{document}